\documentclass[conference, a4paper, 10pt, times]{IEEEtran}
\IEEEoverridecommandlockouts
\usepackage{cite}
\usepackage{amsmath, amssymb, amsfonts, subcaption, booktabs}
\usepackage{algorithmic}
\usepackage{graphicx}
\usepackage{textcomp}
\usepackage{xcolor}
\usepackage{multirow}
\usepackage[linesnumbered,ruled,vlined]{algorithm2e}
\usepackage{amsmath,amsfonts,bm}

\def\eqref#1{equation~\ref{#1}}
\def\1{\bm{1}}

\DeclareMathAlphabet{\mathsfit}{\encodingdefault}{\sfdefault}{m}{sl}
\SetMathAlphabet{\mathsfit}{bold}{\encodingdefault}{\sfdefault}{bx}{n}

\newcommand{\argminD}{\arg\!\min} % AlfC
\usepackage{paralist}
\usepackage{hyperref}
\usepackage{array}
\newtheorem{definition}{Definition}

\def\BibTeX{{\rm B\kern-.05em{\sc i\kern-.025em b}\kern-.08em
    T\kern-.1667em\lower.7ex\hbox{E}\kern-.125emX}}
    
\begin{document}

\title{Rethinking the Trigger-injecting Position in Graph Backdoor Attack}

\author{\IEEEauthorblockN{Jing Xu}
\IEEEauthorblockA{\textit{Delft University of Technology}\\
Delft, Netherlands \\
j.xu-8@tudelft.nl}
\and
\IEEEauthorblockN{Gorka Abad}
\IEEEauthorblockA{\textit{Radboud University}\\
Nijmegen, Netherlands \\
\textit{Ikerlan research centre} \\
Arrasate-Mondragón, Spain \\
abad.gorka@ru.nl}
\and
\IEEEauthorblockN{Stjepan Picek}
\IEEEauthorblockA{\textit{Radboud Univesity}\\
Nijmegen, Netherlands \\
\textit{Delft University of Technology} \\
Delft, Netherlands \\
stjepan.picek@ru.nl}
}

\maketitle

\begin{abstract}
Backdoor attacks have been demonstrated as a security threat for machine learning models. 
Traditional backdoor attacks intend to inject backdoor functionality into the model such that the backdoored model will perform abnormally on inputs with predefined backdoor triggers and still retain state-of-the-art performance on the clean inputs. 
While there are already some works on backdoor attacks on Graph Neural Networks (GNNs), the backdoor trigger in the graph domain is mostly injected into random positions of the sample. There is no work analyzing and explaining the backdoor attack performance when injecting triggers into the most important or least important area in the sample, which we refer to as trigger-injecting strategies MIAS and LIAS, respectively. 
Our results show that, generally, LIAS performs better, and the differences between the LIAS and MIAS performance can be significant. 
Furthermore, we explain these two strategies' similar (better) attack performance through explanation techniques, which results in a further understanding of backdoor attacks in GNNs. 
\end{abstract}


\section{Introduction}
\label{sec:intro}

Graph Neural Networks (GNNs) have demonstrated their superior performance in a variety of applications, such as node classification~\cite{huang2022auc}, graph classification~\cite{errica2019fair}, image classification~\cite{zhou2020graph}, and natural language processing~\cite{zhou2020graph}. However, GNNs are vulnerable to various adversarial attacks, including the backdoor attack. Specifically, a backdoor attack occurs when the adversary deliberately modifies a proportion of the training data by adding the trigger (e.g., subgraph in a graph) to make the model misclassify the samples with the trigger as the target label(s). 
The backdoored GNN model aims to perform normally on benign testing samples. However, if the same trigger used in the training phase is introduced onto a testing sample, the backdoored model exhibits a particular output behavior of the adversary's choosing, such as misclassification into to target label(s)).  
Backdoor attacks have been demonstrated to perform malicious tasks on security-related graph learning services, such as converting the label of a fraud account to benign in a social network~\cite{liu2021pick}. Hence, the backdoor attack is a serious threat to the practical applications of GNNs.

Several works explored the backdoor attacks in GNNs~\cite{xi2021graph, zhang2021backdoor, xu2021explainability}. In these works, one idea of the trigger-injecting position is randomly selecting a subgraph as there is no specific location information in a graph~\cite{wu2020comprehensive}. Another idea to inject the trigger into a graph is to select the subgraph which has high similarity with the trigger graph~\cite{xi2021graph}. 
Moreover, based on the improvement of the explanation techniques in the graph domain,~\cite{xu2021explainability} proposed injecting the trigger into the most important or least important area of the sample. % with the help of the explanation techniques. 
However, that work does not provide any experimental analysis to confirm the assumptions made. 
Also, there is no work so far on using explanation tools to explain
the backdoor attack behavior in the graph domain. 
This work first raises a core question:  

\textit{What is the attack performance when injecting trigger into the most or least important area of the sample?}

To answer this question, we explore the impacts of the backdoor trigger-injecting position from the perspective of the most (MIAS) or least important area of the sample (LIAS). 
Although there is no location information in a graph, we can still locate the most (least) important area in a graph, like in an image, by using some explanation techniques~\cite{ying2019gnnexplainer}. 
As shown in experiments, we demonstrate that the attack performance of LIAS is better, where the difference from MIAS can even be significant.
This observation inspires one further question:

\textit{Can we explain this difference?}

\begin{figure}[!htb]
\centering
     \begin{subfigure}[t]{0.15\textwidth}
         \centering
         \includegraphics[width=\textwidth]{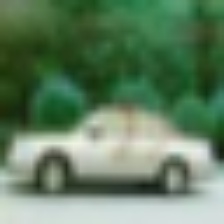}
         \caption{}
         %\caption{Clean image.}
         \label{fig:gradcam_a}
     \end{subfigure}
     \hfill
     \begin{subfigure}[t]{0.15\textwidth}
         \centering
         \includegraphics[width=\textwidth]{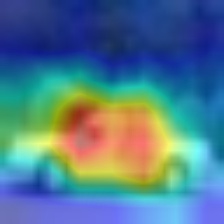}
         \caption{}
         %\caption{Heatmap for the true label on a clean model (predicted as the true label).}
         \label{fig:gradcam_b}
     \end{subfigure}
     \hfill
     \begin{subfigure}[t]{0.15\textwidth}
         \centering
         \includegraphics[width=\textwidth]{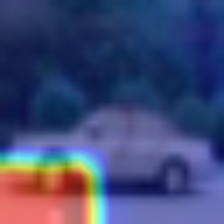}
         \caption{}
         %\caption{Heatmap for target label on a backdoored model (predicted as target label).}
         \label{fig:gradcam_c}
     \end{subfigure}\caption{\label{fig:gradcam}An example of using Grad-CAM to explain a backdoor attack in the image domain. (a) clean image, (b) heatmap of clean image for the true label on the backdoored model (predicted as the true label), (c) heatmap of the poisoned image for the target label on the backdoored model (predicted as the target label).}
\end{figure}

There are already some works on explaining backdoor attacks in the image domain through visualization techniques~\cite{gu2019badnets, xie2020dba}. For example,~\cite{gu2019badnets} plotted the average activations of the backdoored model's last convolutional layer over clean and backdoored images to explain their attack.~\cite{xie2020dba} used the Grad-CAM~\cite{selvaraju2017grad} visualization method to explain the backdoor attack in federated learning. One example of explaining a backdoor attack in the image domain with Grad-CAM is shown in Fig.~\ref{fig:gradcam}. 
Comparing the heatmaps of the clean and poisoned images on the backdoored model, we can clearly understand how the backdoored model recognizes the trigger pattern to achieve the backdoor attack. 
In contrast, applying visualization techniques to explain the backdoor attack behavior in the graph domain is difficult. First, the complexity of the visual representation of a graph is much larger than visualizing an image, especially for large graphs~\cite{hu2015visualizing}. 
Second, visualizing the graph neural networks to explain the backdoor attack is not trivial as it is a time-consuming or even impossible process~\cite{jin2022gnnlens}.

Therefore, in this work, instead of using the visualization method, we explain the difference between two trigger-injecting strategies by computing an evaluation metric. Specifically, we compute the similarity of the predicted mask of the representative features from the backdoored model and the target mask of the representative features from the clean model. %We assume that higher similarity contributes to better attack performance. \todo{why?}
In our experiments, we find that the successfully misclassified samples generally have high similarity while the unsuccessfully misclassified samples have a much lower similarity. 
However, we also find that in one specific case, the high similarity does not lead to a successful attack. We further study this phenomenon and find that the backdoored model trained by MIAS can recognize the original feature pattern in addition to the trigger pattern. 
As there has been an increasing number of studies on trustworthy GNNs~\cite{zhang2022trustworthy, dai2022comprehensive, wang2021confident}, this paper also contributes to the exploration of GNN robustness by investigating the effectiveness of backdoor attacks in GNN models using an explainability tool.

Our work is the first to revisit the trigger-injecting position in graph backdoor attacks and provide a new perspective. Our key contributions are:
\begin{compactenum}
\item We investigate backdoor attacks in GNNs by injecting triggers into the most or least important area of the sample. 
\item We design a novel explanation framework to analyze the causes of the difference between these two strategies.
\item We verify the difference with quantitative analysis (recall score), which helps us further understand the backdoor attack behavior in GNNs.
\item We explore the interaction between the explainability and robustness of GNNs through experiments of two datasets and two GNN models.
\end{compactenum}

\section{Background}
\label{sec:background}

% \subsection{Deep Learning}

% Deep learning (DL) is the automated process of giving a parameterized function $f_\theta(\cdot)$ by $\theta$, mapping an input $\textbf{x} \in X$ to an output $y \in Y$. \footnote{We only focus on supervised learning.} \todo{?}
% For achieving an accurate mapping, $\theta$ is iteratively adjusted by minimizing a loss function, e.g., cross-entropy loss or mean squared error, using a training set $D_{train}$.
% $f_\theta(\cdot)$ can handle different types of data linked to its architecture and design. For instance, convolutional neural networks (CNNs) are used in the image domain, while in the graph domain, graph neural networks are used (GNNs).

% CNNs have achieved superior performance in processing images~\cite{lecun1998gradient, szegedy2015going}. Unlike fully connected neural networks, CNNs use convolutional layers, reducing input size while encoding information. The convolutional layers contain filters that capture the relevant information of the images, such as the colors or shapes. After convolutions,\todo{pooling} the data is commonly processed by fully connected layers, which makes the classification.

\subsection{Graph Neural Networks}

Recently, Graph Neural Networks (GNNs) have achieved significant success in processing non-Euclidean spatial data, which are common in many real-world scenarios~\cite{zhou2020graph}. Unlike traditional neural networks, e.g., Convolutional Neural Networks (CNNs) or Recurrent Neural Networks (RNNs), GNNs work on graph data. 
GNNs take a graph $G=(V,E,X)$ as an input, where $V, E, X$ denote nodes, edges, and node attributes, and learn a representation vector (embedding) for each node $\boldsymbol{v} \in G$, $z_{\boldsymbol{v}}$, or the entire graph, $z_G$. 

In particular, in modern GNNs, the node representation is computed by recursive aggregation and transformation of feature representations of its neighbors. After $k$ iterations of aggregation, a node's representation captures both structure and feature information within its $k$-hop network neighborhood. Formally, the $k$-th layer of a GNN is:
\begin{equation}
x_{\boldsymbol{v}}^{(k)} = AGGREGATION^{(k)}(\left \{ z_{\boldsymbol{v}}^{(k-1)}, \left \{z_{\boldsymbol{u}}^{(k-1)}|\boldsymbol{u}\in \mathcal{N}_{\boldsymbol{v}} \right\} \right \}),
    %z_\boldsymbol{v}^{(k)} = \sigma(z_\boldsymbol{v}^{(k-1)}, agg(\{ z_\boldsymbol{u}^{k-1};\boldsymbol{u}\in \mathcal{N}_\boldsymbol{v}\})), \forall k \in [K].
    \label{equ:gnn_agg}
\end{equation}

\begin{equation}
    z_{\boldsymbol{v}}^{(k)} = TRANSFORMATION^{(k)}(x_{\boldsymbol{v}}^{(k)})
    \label{equ:gnn_trans},
\end{equation}
where $z_{\boldsymbol{v}}^{(k)}$ is the representation of node $\boldsymbol{v}$ computed in the $k$-th iteration. $\mathcal{N}_{\boldsymbol{v}}$ are 1-hop neighbors of node $\boldsymbol{v}$, and the $AGGREGATION(\cdot)$ is an aggregation function that can vary for different GNN models. $z_{\boldsymbol{v}}^{(0)}$ is initialized as node feature. The $TRANSFORMATION(\cdot)$ function consists of a learnable weight matrix and activation function. 
For the node classification task, the node representation $z_{\boldsymbol{v}}$ is used for prediction. 
In this paper, we investigate the node classification task. 
% For the graph classification task, the READOUT function pools the node representations for a graph-level representation $z_G$:
% \begin{equation}
%     z_G = READOUT({z_{\boldsymbol{v}};v \in V}).
%     \label{eqn:2.2-2}
% \end{equation}
% READOUT can be a simple permutation invariant function such as summation or a more sophisticated graph-level pooling function~\cite{DBLP:conf/nips/YingY0RHL18,DBLP:conf/aaai/ZhangCNC18}.
Moreover, we focus on two representation models of this family, which differ in one of the above two steps: aggregation and transformation. In the following, we briefly describe these models and their differences.

\noindent \textbf{Graph Convolutional Networks (GCN)~\cite{kipf2017semi}.} Let $d_{\boldsymbol{v}}$ denotes the degree of node $\boldsymbol{v}$. The aggregation operation in GCN is then given as:
%\todo{This equation (and subsequent ones) is not referenced. Consider removing the number}
\begin{equation*}
    x_{\boldsymbol{v}}^{(k)} \leftarrow \sum_{u \in \mathcal{N}_{\boldsymbol{v}}\bigcup {\boldsymbol{v}}} \frac{1}{\sqrt{d_{\boldsymbol{v}}d_{\boldsymbol{u}}}}z_{\boldsymbol{u}}^{(k-1)}.
\end{equation*}
GCN performs a non-linear transformation over the aggregated features to compute the node representation at layer $k$:
\begin{equation*}
    z_{\boldsymbol{v}}^{(k)} \leftarrow ReLU(x_{\boldsymbol{v}}^{(k)}W^{(k)}).
\end{equation*}

% \noindent \textbf{Graph Isomorphism Network (GIN)~\cite{xu2018how}.} GIN is developed as powerful as the Weisfeiler-Lehman graph isomorphism test~\cite{weisfeiler1968reduction}. Let $\epsilon$ be a learnable parameter or a fixed scalar. Then, the aggregation operation in GIN is given as:
% \begin{equation}
%     x_{\boldsymbol{v}}^{(k)} \leftarrow (1+\epsilon^{(k)})\cdot z_{\boldsymbol{v}}^{(k-1)} + \sum_{\boldsymbol{u} \in \mathcal{N}_{\boldsymbol{v}}}z_{\boldsymbol{u}}^{(k-1)}
% \end{equation}

% In GIN, a multi-layer perceptrons (MLPs) is used as the transformation operation:
% \begin{equation}
%     z_{\boldsymbol{v}} \leftarrow MLP^{(k)}(x_{\boldsymbol{v}}^{(k)})
% \end{equation}

\noindent \textbf{Graph Attention Networks (GAT)~\cite{velickovic2018graph}.} In addition to the standard neighbor aggregation scheme mentioned above in ~\eqref{equ:gnn_agg} and~\eqref{equ:gnn_trans}, there are other non-standard neighbor aggregation schemes, e.g., weighted average via attention in GAT. 
Specifically, given a shared attention mechanism $a$, attention coefficients can be computed by:
\begin{equation}
    e_{\boldsymbol{v}\boldsymbol{u}} = a(Wz_{\boldsymbol{v}}^{(k-1)}, Wz_{\boldsymbol{u}}^{(k-1)})
\end{equation}
that indicate the importance of node $\boldsymbol{u}$'s features to node $\boldsymbol{v}$. Then, the normalized coefficients can be computed by using the softmax function:
\begin{equation}
    \alpha _{\boldsymbol{v}\boldsymbol{u}} = softmax_{\boldsymbol{u}}(e_{\boldsymbol{v}\boldsymbol{u}}).
\end{equation}
Finally, the next-level feature representation of node $\boldsymbol{v}$ is:
\begin{equation}
    z_{\boldsymbol{v}}^{(k)} = \sigma \left ( \frac{1}{P}\sum_{p=1}^{P}\sum_{\boldsymbol{u}\in\mathcal{N}_{\boldsymbol{v}}}\alpha _{\boldsymbol{v}\boldsymbol{u}}^pW^pz_{\boldsymbol{u}}^{(k-1)} \right ),
\end{equation}
where $\alpha _{\boldsymbol{v}\boldsymbol{u}}^p$ are the normalized coefficients computed by the $p$-th attention mechanism $a^p$ and $W^p$ is the corresponding input linear transformation's weight matrix.

\subsection{Backdoor Attacks}

%Backdoor attacks are a prominent threat in DL~\cite{li2022backdoor}.\todo{DL has not been previously introduced}
Backdoors are training time attacks that aim to achieve misclassification at the testing phase for trigger-embedded samples while working correctly on clean inputs. 
%More precisely, backdoor attacks modify the training set to include a trigger, thus generating a set of poisoned samples $\hat{\textbf{x}}\in D_{bk}$ while $\hat{\textbf{x}}$ is assigned a target label $\hat{y}$. The attacker chooses the amount of poisoned data for training, controlled by $\epsilon = \frac{|D_{bk}|}{|D_{bk}| + |D_{train}|}$. Choosing a $\epsilon$ is crucial for finding a good balance between the backdoor and clean tasks. A larger $\epsilon$ value will lead to a more powerful backdoor but a drop in the performance of the main task and vice-versa.
Several studies showed that GNNs are also vulnerable to backdoor attacks. Similar to the backdoor attack in CNNs, the backdoor attack in GNNs can be implemented by poisoning the training data with a trigger, which can be a subgraph with/without features~\cite{zhang2021backdoor,xi2021graph} or a subset of node features~\cite{xu2021explainability}. After training the GNN model with the trigger-embedded data, the backdoored GNN would predict the test example injected with a trigger as the pre-defined target label.

\subsection{Explainability of GNNs}

Recently, several explainability techniques in GNNs have been proposed, such as XGNN~\cite{yuan2020xgnn}, GNNExplainer~\cite{ying2019gnnexplainer}, PGExplainer~\cite{luo2020parameterized}, and SubgraphX~\cite{yuan2021explainability}. These methods are developed from different angles and provide different levels of explanations. 

GNNExplainer is the model-agnostic approach for providing explanations on any GNN-based model's predictions. Given a trained GNN model and its prediction(s), GNNExplainer returns an explanation in the form of a small subgraph of the input graph, with a small subset of node features that contribute most to the final model prediction(s). 
In this paper, we focus on the GNNExplainer method as it can explain predictions of any GNN on any graph-based machine learning task without requiring modification of the underlying GNN architecture or re-training.  %because this work is supposed to further explain the results in~\cite{xu2021explainability}. 

\section{Threat Model}
\label{sec:threat}

We consider a \emph{gray-box} threat model assuming the attacker can freely modify a small portion of the training dataset. 
Since the explanation masks in GNNExplainer are generated through gradients of the GNN model, the attacker also has knowledge of the gradient information of the target model on the chosen training dataset. 
We also assume the attacker performs a \emph{dirty-label} backdoor attack, where the poisoned samples' labels are changed to the target label. Although this kind of attack is weaker than \emph{clean-label} backdoor attacks~\cite{clean-label-backdoor-attacks}, where the labels remain unaltered, dirty label attacks are the most common in the literature~\cite{xi2021graph, zhang2021backdoor, xu2021explainability}. 
%Given a pre-trained clean GNN model, we train it over the poisoned training dataset to obtain the backdoored model. 
The attacker's goal is to inject a backdoor in the given pre-trained clean GNN model through training over the poisoned training dataset, which achieves misclassification under the presence of a trigger while maintaining clean high accuracy on the original task. 
This threat model is realistic in real-world settings. For example, if the training dataset is collected from public users, the adversary can provide trigger-embedded training data to implement the backdoor attack. 

\section{Methodology}
\label{sec:methodology}

\subsection{General Framework}

%\todo{I still did not see explanation of LIA or MIA, maybe put in threat model}

As stated before, we aim to discover if and how the explainability techniques in GNNs help improve the performance of backdoor attacks. 
Here, we focus on utilizing the feature-trigger backdoor attack from~\cite{xu2021explainability} for the node classification task. 
The trigger used in the backdoor attacks in our paper is defined as:
\begin{definition}[Trigger] In our backdoor attacks, the trigger is a specific feature pattern that is created by modifying the value of a subset of a node's features.
\end{definition}

Generally, two steps are conducted to implement backdoor attacks using explainability techniques:

(1) We apply an explainability technique (i.e., GNNExplainer) on a pre-trained clean GNN model to implement backdoor attacks based on two trigger-injecting strategies defined below. 
\begin{definition}[The Most/Least Representative Features]Through applying the GNNExplainer on the pre-trained clean GNN model on the target node, we can obtain the original importance order of the node features. Based on the importance order information, we can locate the most or least representative features. 
\end{definition}

\begin{definition}[Most Important Area Strategy (MIAS)] We select the most representative features of the target node and inject the feature trigger into the corresponding dimensions.
\end{definition}

\begin{definition}[Least Important Area Strategy (LIAS)] We select the least representative features of the target node and inject the feature trigger into the corresponding dimensions.
\end{definition}

We then compare the attack performance based on these two strategies, including the attack success rate and clean accuracy drop. 

\begin{figure}[!htb]
\centering
     \begin{subfigure}[b]{0.48\textwidth}
         \centering
         \includegraphics[width=\textwidth, page=1]{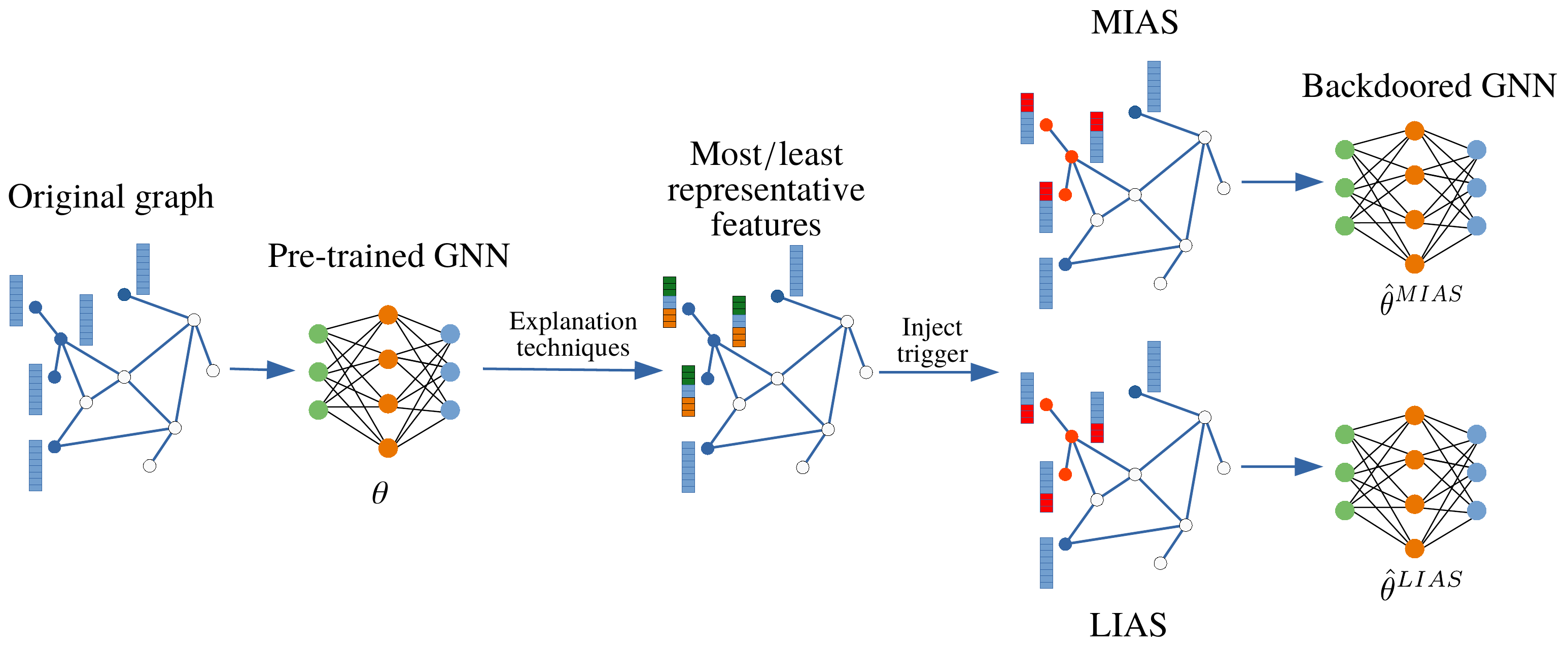}
         \caption{Backdoor attacks based on two strategies.}
         \label{fig:framework_graph_a}
     \end{subfigure}
     \vfill
     \begin{subfigure}[b]{0.48\textwidth}
         \centering
         \includegraphics[width=\textwidth, page=2]{figs/framework_graph.pdf}
         \caption{Procedure of explaining two attack strategies.}
         \label{fig:framework_graph_b}
     \end{subfigure}
\caption{\label{fig:framework_graph}An illustration of backdoor attack and explanation framework.}
\end{figure}

(2) Next, we try to explain the attack performance of these two strategies by again applying the explainability techniques on the backdoored model over the poisoned testing dataset. 
As a result, we can obtain the new importance order of the node features, which is used to compute the similarity with the original feature importance order.
The proposed framework is presented in Fig.~\ref{fig:framework_graph}. 

\subsection{Explanation Design}

%Specifically,  We utilize one of the most powerful explanation techniques in GNNs - GNNExplainer to implement backdoor attacks of MIAS and LIA strategies, i.e., we find the most/least representative features of the target node and inject the feature trigger into these specific feature dimensions. 
 
The detailed process of generating poisoned training dataset and target masks is presented in Algorithm~\ref{alg:generate_poisoned_dataset}. $EXP(\cdot)$ is the applied GNN explanation technique, i.e., GNNExplainer, and $s$ is the trigger-injecting strategies, i.e., MIAS or LIAS. The algorithm first samples a subset from the original training dataset with a poisoning rate $r$ (line $2$). For each sampled node, the algorithm will compute the corresponding feature order to determine the trigger-injecting location for MIAS and LIAS. Meanwhile, the label of the poisoned training dataset will change to the target label. 
The trigger size $n$ is the number of the features in the feature trigger, which means $n$ node features will be modified.  
The poisoned testing dataset is obtained by injecting a trigger (following the same strategy as the poisoned training dataset) into the samples %which\todo{?} have different labels from the target label 
and changing their labels to the target label. Finally, based on the order of representative features, we can generate a target mask for each node in the poisoned testing dataset (line $17$). The target mask has the same shape as the node feature vector, and the most (least) $n$ important features are masked in while other features are masked out. 
To evaluate whether the backdoored model can recognize the trigger pattern precisely, the number of features to be masked in is set to be $n$. 
The definition of the target mask is as follows: 
\begin{definition}[Target Mask] The target mask is a boolean tensor that indicates which $n$ features contribute more to the final prediction from the \textit{pre-trained clean model $\theta$} for the target node compared to other features.
\end{definition}
% The target mask indicates for the target node which features contribute more or less to the final prediction from the \textbf{pre-trained clean model $\theta$}.  

\begin{algorithm}
\small
\SetAlgoLined
\caption{Generate Poisoned Training Dataset and Target Masks} \label{alg:generate_poisoned_dataset}
\SetKwInput{KwInput}{Input}
\SetKwInput{KwOutput}{Output}
\DontPrintSemicolon
    \KwInput{
    Pre-trained clean GNN model $\theta$, Training set $D_{train}$, Testing set $D_{test}$, Trigger-injecting strategy $s\in \left \{ MIAS, LIAS\right \}$, Target label $y_t \in \left [0, C\right )$\\}
    \KwOutput{
    Poisoned training dataset $\hat{D}_{train}^s$, Poisoned testing dataset $\hat{D}_{test}^s$, Target masks $M_t^s$\\
    }
        /* Sampling Training Dataset to Inject Trigger */ \;
        $\hat{D}_{train}^s \leftarrow sample(D_{train}, r, y \neq y_t)$\;
        \ForEach{$\left \{x, y \right \} \in \hat{D}_{train}^s$}
        {
            /* Computing Order of Representative Features */ \;
            $feature\_order = EXP(\theta, x, y)$ \;
            $\hat{x}^s = Inject\_Trigger(x, feature\_order, s)$ \;
            $\hat{y}^s = y_t$\;
            %$D_{train}^{poi} = D_{train}^{poi} \cup \left\{x_{poi}, y_{poi} \right\}$\;
        }
        $\hat{D}_{test}^s \leftarrow D_{test}[\setminus y_t]$ \;
        $M_t^s \leftarrow \emptyset $ \;
        \ForEach{$\left \{x, y \right \} \in \hat{D}_{test}^s$}
        {
            /* Computing Order of Representative Features */ \;
            $feature\_order = EXP(\theta, x, y)$ \;
            $\hat{x}^s = Inject\_Trigger(x, feature\_order, s)$ \;
            $\hat{y}^s = y_t$\;
            /* Generating Target Mask */ \;
            $m_i^s = Get\_Mask(feature\_order, s)$ \;
            $M_t^s = M_t^s \cup m_i^s$
        }
    \textbf{return} $\hat{D}_{train}^s, \hat{D}_{test}^s, M_t^s$
\end{algorithm}

\begin{algorithm}
\small
\SetAlgoLined
\caption{Train Backdoored GNN Models and Generate Predicted Masks}
\label{alg:train_bkd}
\SetKwInput{KwInput}{Input}
\SetKwInput{KwOutput}{Output}
\DontPrintSemicolon
    \KwInput{
    Pre-trained clean GNN model $\theta$, Training set $D_{train}$, Poisoned training dataset $\hat{D}_{train}$, Poisoned testing dataset $\hat{D}_{test}$, Trigger-injecting strategy $s\in \left \{ MIAS, LIAS\right \}$\\}
    \KwOutput{
    Backdoored GNN model $\hat{\theta}^s$, Predicted Masks $M_p^s$\\
    }
    /* Training Backdoored Models */ \;
    /* $\left \{ x, y \right \} \in D_{train}$, $\left \{ \hat{x}^s, \hat{y}^s \right \} \in \hat{D}_{train}^s$ */\;
    $\hat{\theta}^s= \argminD_{\theta}(\sum_{i}L(x_i, y_i;\theta)+\sum_{i}L(\hat{x}_i^s,\hat{y}_i^s;\theta))$ \;
    $M_p^s \leftarrow \emptyset $ \;
        \ForEach{$\left \{\hat{x}^s, \hat{y}^s \right \} \in \hat{D}_{test}^s$}
        {
            /* Getting Predictive Mask */ \;
            $feature\_order = EXP(\hat{\theta}^s, \hat{x}^s, \hat{y}^s)$ \;
            $m_i^s = Get\_Mask(feature\_order)$ \;
            $M_p^s = M_p^s \cup m_i^s$
        }
    \textbf{return} $\hat{\theta}^s, M_p^s$
\end{algorithm}

Once the poisoned training dataset is generated, we can obtain the backdoored models $\hat{\theta}^s$ by retraining the clean model $\theta$ with the backdoored training dataset.~\footnote{In this work, we combine the original training dataset and the poisoned training dataset as the backdoored training dataset.} The process of training the backdoored models and obtaining predicted masks is shown in Algorithm~\ref{alg:train_bkd}. 
To analyze the impact of injecting trigger into the most/least important part of the node features on the attack performance, we compare the attack performance of $\hat{\theta}^{MIAS}$ and $\hat{\theta}^{LIAS}$, including the attack success rate and clean accuracy drop. 
Finally, for the poisoned testing dataset, which we used to calculate the attack success rate, we again utilize the GNNExplainer to obtain the new feature importance order for each node on the backdoored GNN model $\hat{\theta}^{MIAS}$ or $\hat{\theta}^{LIAS}$ (line $7$). 
The new feature importance order is used to generate the predicted mask. 
The definition of the predicted mask is as follows:
\begin{definition}[Predicted Mask] The predicted mask is a boolean tensor which indicates which $n$ features contribute more to the final prediction from the \textit{backdoored GNN model $\hat{\theta}$} for the target node compared to other features.
\end{definition}
 
Combining the target masks we get in Algorithm~\ref{alg:generate_poisoned_dataset}, we can compute the similarity between the ordering of the new representative features and the old ones by calculating the recall score of the target mask and the predicted mask:
\begin{equation}
\begin{split}
    & RS_i^s = \frac{TP(M_{t, i}^s, M_{p, i}^s)}{TP(M_{t, i}^s, M_{p, i}^s) + FN(M_{t, i}^s, M_{p, i}^s)}, i \in N \\
    & M_{t, i}^s (M_{p, i}^s) = [0,\cdots, 1, \cdots,1, \cdots, 0],
\end{split}
\end{equation}
where $RS_i^s$ is the recall score of the $i$th poisoned testing sample with $s$ strategy, $M_{t, i}^s$ and $M_{p, i}^s$ is the target mask, and predictive mask of the $i$th poisoned testing sample, $TP$ and $FN$ is the true positive and false negative rate of these two masks, respectively, and $N$ is the number of the poisoned testing dataset. 
We assume that higher similarity indicates that the backdoored model can better recognize the trigger pattern, contributing to better attack performance. 

\section{Experimental Results}
\label{sec:experiments}

% In this section, we conduct empirical studies to answer the following research questions:
% \begin{quote}
% \textbf{Q1}: Will MIA strategy result in better attack performance than LIA?\\
% \textbf{Q2}: Why MIA (LIA) strategy has better attack performance than LIA (MIA)?
% \end{quote}
% \todo{???}

\subsection{Experimental Setting}
We implemented the backdoor attack on the node classification task using the PyTorch framework. All experiments were run on a server with $2$ Intel Xeon CPUs, $1$ NVIDIA 1080 Ti GPU with $32$GB RAM, and Ubuntu $20.04$ LTS OS. Each experiment was repeated $10$ times to obtain the average result. 

\noindent \textbf{Dataset.} 
For our experiments, we use two publicly available real-world datasets for the node classification task: Cora~\cite{sen2008collective} and CiteSeer~\cite{sen2008collective}. These two datasets are citation networks in which each publication is described by a binary-valued word vector indicating the absence/presence of the corresponding word in the collection of $1,433$ and $3,703$ unique words, respectively. 
%we use two publicly available real-world graph datasets. (i) Mutagenicity~\cite{morris2020tudataset} - molecular structure graphs of mutagen and nonmutagen; (ii) REDDIT-BINARY~\cite{yanardag2015deep} - a dataset consisting of graphs corresponding to online discussions on Reddit, for the graph classification task. 

%\noindent \textbf{Dataset splits and parameter setting.} 
%For each graph classification dataset, we sample $80\%$ of the graphs as the original training dataset and treat the remaining graphs as the original testing dataset. 
%Among the original training dataset, we randomly sample $\eta=0.05$ fraction of graphs to inject the trigger and relabel them with the target label, called the backdoored training dataset. The trigger size and trigger density are set to be the $\gamma=0.2$ fraction of the graph dataset's average number of nodes and $\rho=0.8$, respectively. And we use Erdős-Rényi (ER) model~\cite{gilbert1959random} to generate the trigger similar to~\cite{xu2021explainability}. 
For each node classification dataset, we split $20\%$ of the total nodes as the original training dataset (labeled), and the rest of the nodes are treated as the original testing dataset. To generate the backdoored training dataset, we sample $10\%$ of the original training dataset to inject the feature trigger and relabel these nodes with the target label. The trigger size is set to $5\%$ of the total number of node feature dimensions. We set these parameters as they provided the best results after conducting a tuning phase. 

\noindent \textbf{Models and training.}
We use the popular GAT~\cite{velickovic2018graph} and GCN~\cite{kipf2017semi} models, as these two methods are commonly-used GNN models for the node classification task. 
We train the clean and backdoored GNN models with a learning rate of $0.005$ and use Adam as the optimizer. 

\noindent \textbf{Attack evaluation metrics.} To compare the attack performance of MIAS and LIAS, we utilize two commonly used backdoor attack evaluation metrics: 
\begin{compactenum}
    \item \textbf{Attacks Success Rate} (ASR): measures the backdoor performance of the model on a fully poisoned dataset $\hat{D}$.  
    It is computed as $ASR = \frac{\sum_{i=1}^{N}\mathbb{I}({\hat{\theta}}(\hat{x_i})=y_t)}{N}$ where ${\hat{\theta}}$ is the poisoned model, $\hat{x_i}$ is a poisoned input, $\hat{x_i} \in \hat{D}$, $y_t$ is the target class, and $\mathbb{I}$ is an indicator function.
    \item \textbf{Clean Accuracy Drop} (CAD): measures the effect of the backdoor attack on the original task. It is calculated by comparing the performance of the poisoned and clean models on a clean holdout testing set. The accuracy drop should generally be small to keep the attack stealthy. 
\end{compactenum}

\subsection{Results and Analysis}

\begin{figure}[!htb]
\centering
     \begin{subfigure}[b]{0.48\textwidth}
         \centering
         \includegraphics[width=\textwidth]{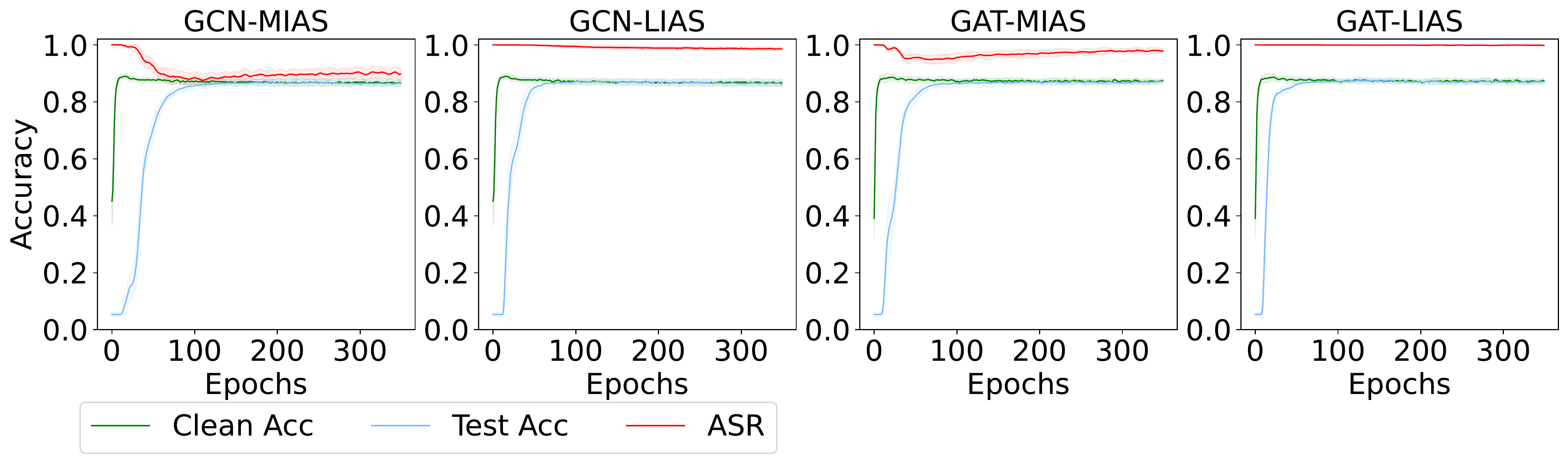}
         \caption{Cora.}
         \label{fig:graph_results_a}
     \end{subfigure}
     \vfill
     \begin{subfigure}[b]{0.48\textwidth}
         \centering
         \includegraphics[width=\textwidth]{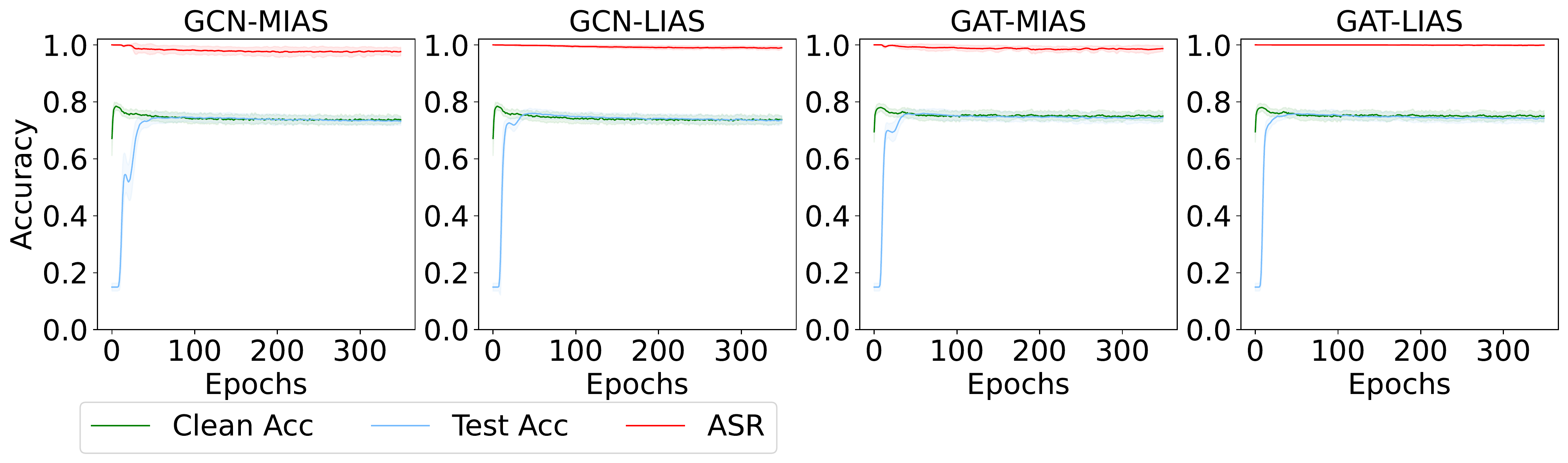}
         \caption{CiteSeer.}
         \label{fig:graph_results_b}
     \end{subfigure}
\caption{\label{fig:graph_results}Backdoor attack results of two trigger-injecting strategies.}
\end{figure}

\noindent \textbf{Results.} The backdoor attack results on two graph datasets based on two models and two trigger-injecting strategies are shown in Fig.~\ref{fig:graph_results}. In particular, the ASR and CAD of two GNN models on two datasets are presented in Table~\ref{Table:bkd_graph_domain}. We can observe that both strategies can achieve a high attack success rate, i.e., more than $97\%$, except GCN on the Cora dataset with MIAS. %\todo{position irrelevant?}. 
In addition, in most cases, the ASR of LIAS is slightly higher, around $1\%$, than that of MIAS. However, for the GCN model on the Cora dataset, the ASR of LIAS is significantly higher: more than $8\%$, than the MIAS. 
We can also see that the CAD for all datasets and models is unnoticeable, and the difference between the two strategies over CAD is negligible. 

% \begin{table}[!htpb]
%  \caption{Backdoor attack performance on the node classification task. \todo{Are this the best results or the average? If the avg, could we also show the std error?}}
% \begin{center}
% \begin{tabular}{cccccc} 
% \hline
%  \multirow{2}{*}{Dataset} & \multirow{2}{*}{Model} & \multicolumn{2}{c}{MIAS} & \multicolumn{2}{c}{LIAS} \\
%  \cmidrule(lr){3-4} \cmidrule(lr){5-6}
%   & & {ASR(\%)} & {CAD(\%)} & {ASR(\%)} & {CAD(\%)}\\
% \hline
% \multirow{2}{*}{Cora} & GCN & $90.08$ & $0.32$ & $98.65$ & $0.27$ \\
% \cline{2-6}
%  & GAT & $97.91$ & $0.34$ & $99.89$ & $0.27$ \\
% \hline
% \multirow{2}{*}{CiteSeer} & GCN & $97.70$ & $0.32$ & $98.96$ & $0.15$ \\
% \cline{2-6}
%  & GAT & $98.54$ & $0.71$ & $99.88$ & $0.80$ \\
% \hline
%  \end{tabular}
% \label{Table:bkd_graph_domain}
% \end{center}
% \end{table}

% \begin{table}
% \begin{center}
% \begin{tabular}{|l||*{5}{c|}}\hline
% \backslashbox{Room}{Date}
% &\makebox[3em]{5/31}&\makebox[3em]{6/1}&\makebox[3em]{6/2}
% &\makebox[3em]{6/3}&\makebox[3em]{6/4}\\\hline\hline
% Meeting Room &&&&&\\\hline
% Auditorium &&&&&\\\hline
% Seminar Room &&&&&\\\hline
% \end{tabular}
% \end{center}
% \end{table}

\begin{table*}[!htpb]
 \caption{Backdoor attack performance of MIAS and LIAS (SD: standard deviation).}
\begin{center}
\begin{tabular}{>{\centering\arraybackslash}m{0.1\textwidth}>{\centering\arraybackslash}m{0.15\textwidth}>{\centering\arraybackslash}m{0.15\textwidth}>{\centering\arraybackslash}m{0.15\textwidth}>{\centering\arraybackslash}m{0.15\textwidth}}
\toprule
\multicolumn{5}{c}{MIAS} \\
\midrule
 \multirow{2}{*}{Dataset} & \multicolumn{2}{c}{GCN} & \multicolumn{2}{c}{GAT} \\
\cmidrule(lr){2-3} \cmidrule(lr){4-5}
 & {ASR $\pm$ SD} & {CAD $\pm$ SD} & {ASR $\pm$ SD} & {CAD $\pm$ SD}\\
\midrule
{Cora} & $90.08\% \pm 0.29\%$ & $0.32\% \pm 0.19\%$ & $97.91\% \pm 0.12\%$ & $0.34\% \pm 0.24\%$ \\
{CiteSeer} & $97.70\% \pm 0.10\%$ & $0.32\% \pm 0.17\%$ & $98.54\% \pm 0.09\%$ & $0.71\% \pm 0.20\%$ \\
\bottomrule
\bottomrule
\multicolumn{5}{c}{LIAS} \\ 
\midrule
 \multirow{2}{*}{Dataset} & \multicolumn{2}{c}{GCN} & \multicolumn{2}{c}{GAT} \\
\cmidrule(lr){2-3} \cmidrule(lr){4-5}
 & {ASR $\pm$ SD} & {CAD $\pm$ SD} & {ASR $\pm$ SD} & {CAD $\pm$ SD}\\
\midrule
{Cora} & $98.65\% \pm 0.06\%$ & $0.27\% \pm 0.21\%$ & $99.89\% \pm 0.03\%$ & $0.27\% \pm 0.21\%$ \\
{CiteSeer} & $98.96\% \pm 0.07\%$ & $0.15\% \pm 0.18\%$ & $99.88\% \pm 0.03\%$ & $0.80\% \pm 0.17\%$ \\
\bottomrule
 \end{tabular}

\label{Table:bkd_graph_domain}
\end{center}
\end{table*}

\noindent \textbf{Analysis.} Next, we investigate the reason why the backdoor attack performance of the LIAS is somewhat higher or significantly higher (for the GCN model on the Cora dataset) than the MIAS. As mentioned in Section~\ref{sec:methodology}, we evaluate the similarity between the ordering of the new representative features and the old ones by calculating the recall score of the target mask and the predicted mask.  
% \iffalse
% In specific, we generate a target mask based on the original representative features where the most (least) $n$ features are masked in and other features are masked out. 
% Here, the number of features to be masked in, i.e., $n$, is set to be the number of features in the feature trigger. 
% And given the backdoored GNN model, we can also generate a predict mask based on the new representative features. We evaluate the similarity between the target mask and predict mask using recall score:

\begin{figure}[!htb]
\centering
     \begin{subfigure}[b]{0.48\textwidth}
         \centering
         \includegraphics[width=\textwidth]{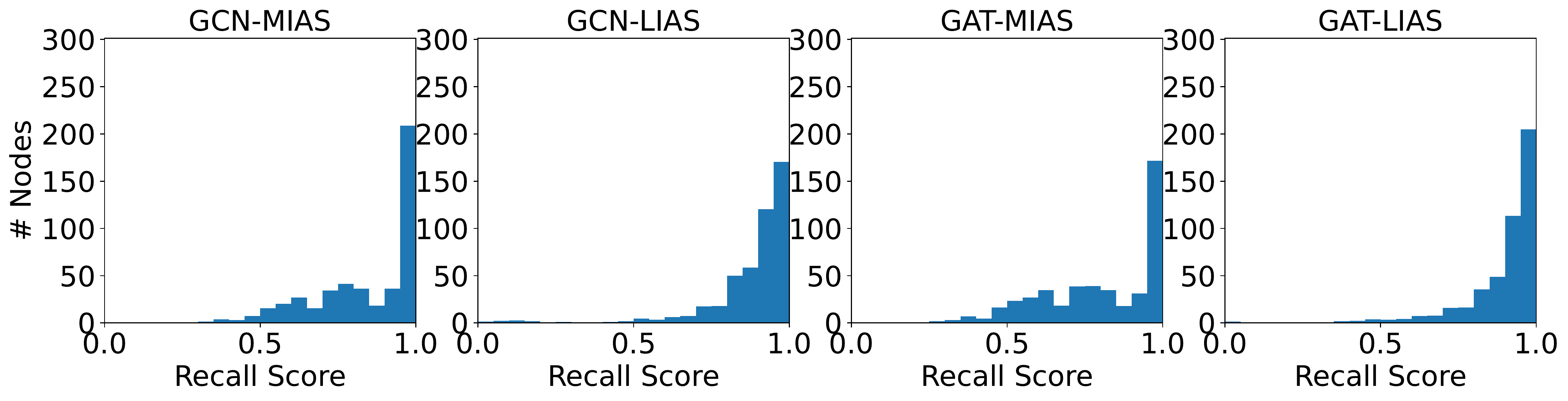}
         \caption{Cora.}
         \label{fig:graph_evaluate_a}
     \end{subfigure}
     \vfill
     \begin{subfigure}[b]{0.48\textwidth}
         \centering
         \includegraphics[width=\textwidth]{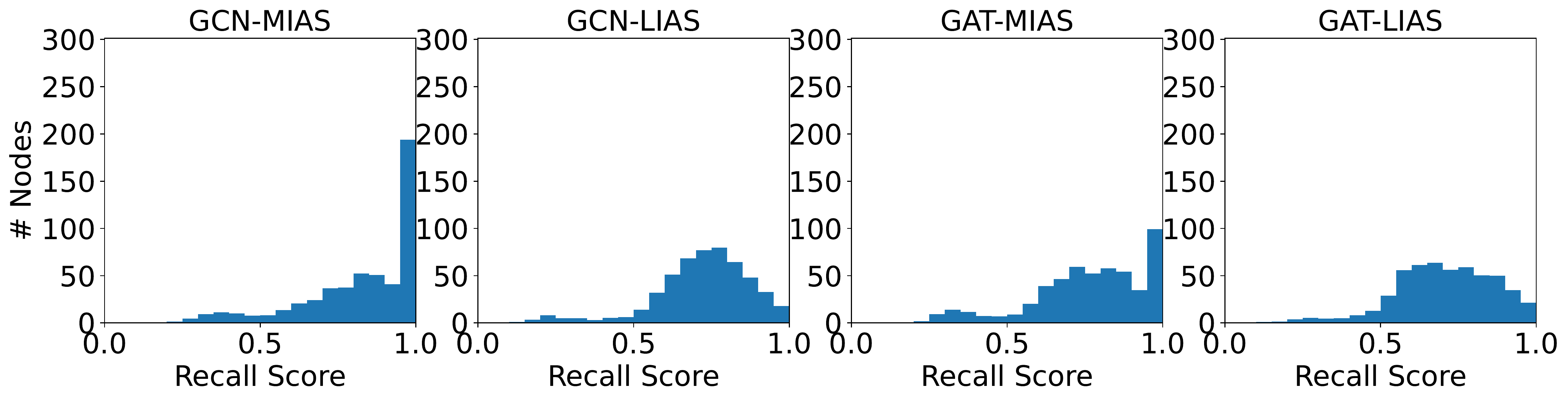}
         \caption{CiteSeer.}
         \label{fig:graph_evaluate_b}
     \end{subfigure}
\caption{\label{fig:graph_evaluate}Histogram of recall scores over the poisoned testing dataset.}
\end{figure}

\begin{figure}[!htb]
\centering
     \begin{subfigure}[b]{0.48\textwidth}
         \centering
         \includegraphics[width=\textwidth]{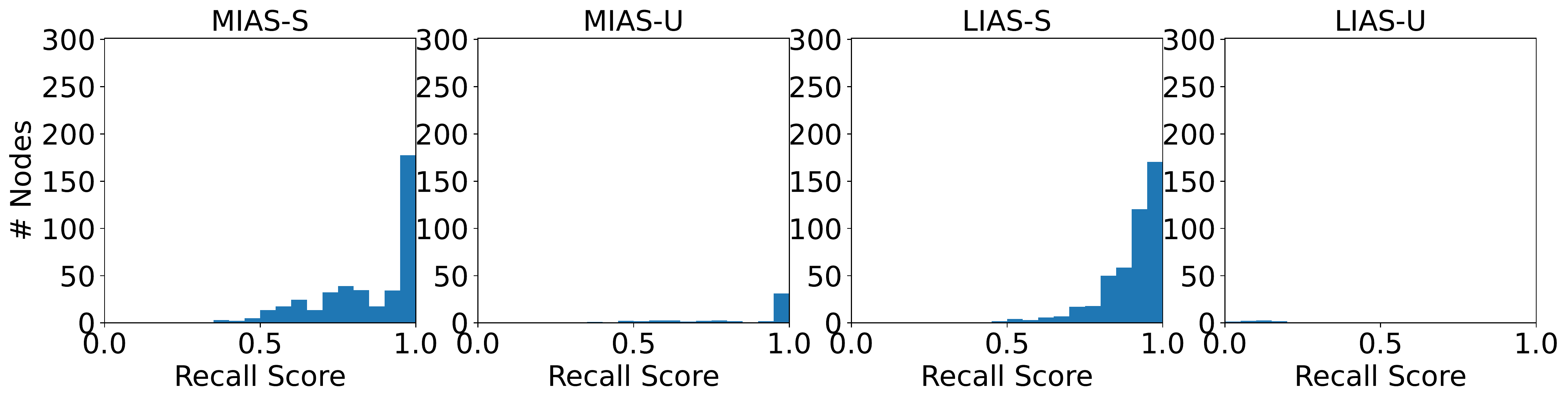}
         \caption{Cora-GCN.}
         \label{fig:cora-gcn}
     \end{subfigure}
     \vfill
     \begin{subfigure}[b]{0.48\textwidth}
         \centering
         \includegraphics[width=\textwidth]{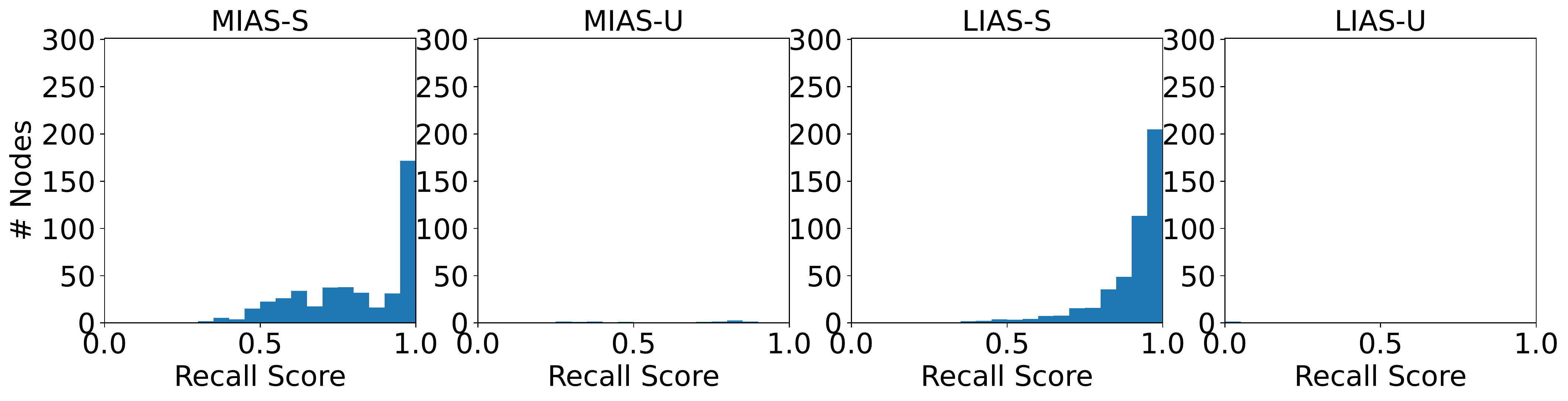}
         \caption{Cora-GAT.}
         \label{fig:cora-gat}
     \end{subfigure}
     \vfill
     \begin{subfigure}[b]{0.48\textwidth}
         \centering
         \includegraphics[width=\textwidth]{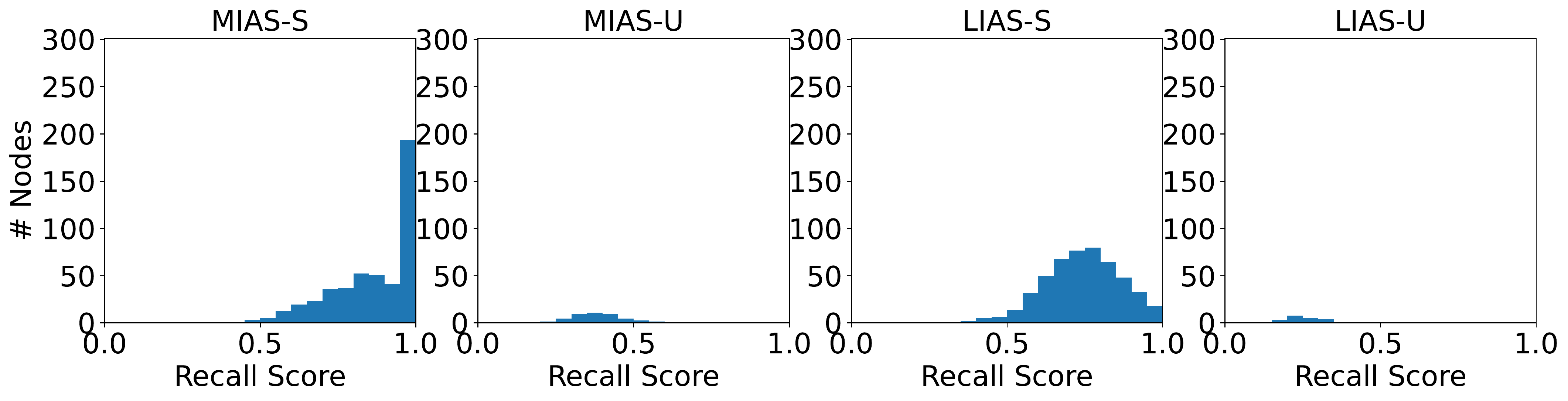}
         \caption{CiteSeer-GCN.}
         \label{fig:citeseer-gcn}
     \end{subfigure}
     \vfill
     \begin{subfigure}[b]{0.48\textwidth}
         \centering
         \includegraphics[width=\textwidth]{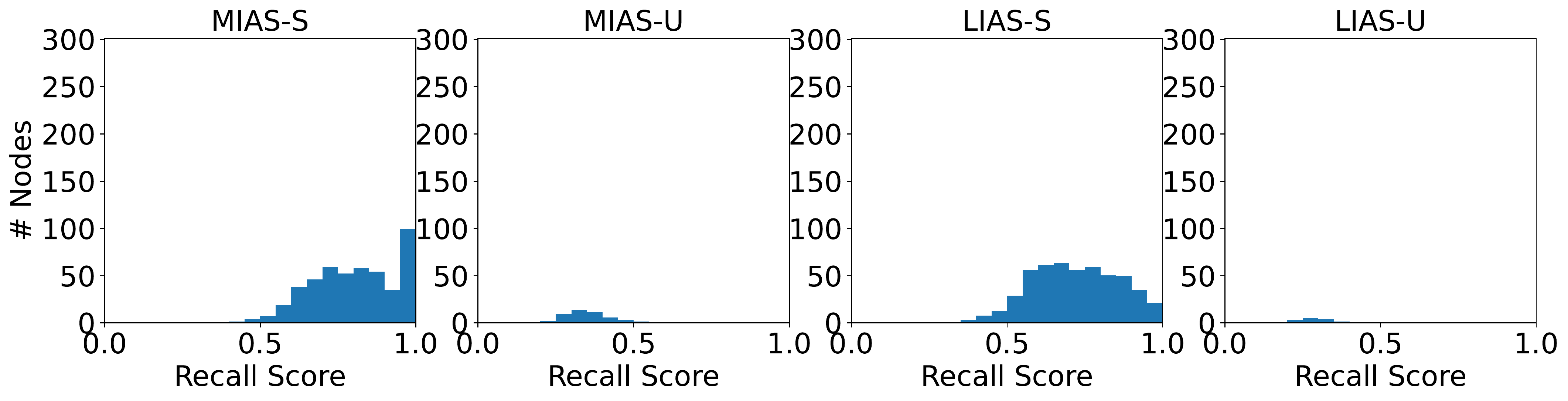}
         \caption{CiteSeer-GAT.}
         \label{fig:citeseer-gat}
     \end{subfigure}
\caption{\label{fig:graph_evaluate_split}Histogram of recall scores over two parts of the poisoned testing dataset ($S$ means the nodes are successfully misclassified into the target label, $U$ means not).}
\end{figure}

% \begin{equation}
% \begin{split}
%     & RS^i = \frac{TP(M_t^i, M_p^i)}{TP(M_t^i, M_p^i) + FN(M_t^i, M_p^i)}, i \in N \\
%     & M_t^i (M_p^i) = [0,\cdots, 1, \cdots,1, \cdots, 0]
% \end{split}
% \end{equation}
% where $RS^i$ is the recall score of the $i$th poisoned testing sample, $M_t^i$ and $M_p^i$ is the target mask and predictive mask of the $i$th poisoned testing sample, $TP$ and $FN$ is the true positive and false negative rate of these two masks, respectively, $N$ is the number of the poisoned testing dataset. 
% \fi
The histogram of recall scores over the poisoned testing dataset of all datasets and models is shown in Fig.~\ref{fig:graph_evaluate}. We can observe that most poisoned testing samples have a recall score of more than $0.5$ in both MIAS and LIAS, which results in a high attack success rate for both strategies. 
To further investigate the slight advantage of the LIAS over the MIAS, we split the poisoned testing samples into two parts, one is misclassified into the target class successfully, and the other one is not, and compute the recall scores for these two parts, as shown in Fig.~\ref{fig:graph_evaluate_split}. 
We notice that, generally, the successfully misclassified nodes have significantly higher recall scores than those not misclassified into the target class. This phenomenon is consistent with the assumption mentioned in Section~\ref{sec:methodology}, i.e., the higher similarity between the ordering of the new representative features and that of the original ones indicates that the backdoored model can recognize the trigger pattern better. 
When comparing the second column and the last column of Fig.~\ref{fig:cora-gat},~\ref{fig:citeseer-gcn}, and~\ref{fig:citeseer-gat}, we also see that LIAS has fewer nodes with low recall score than MIAS, which we believe is the reason of higher ASR of LIAS than MIAS. 

\begin{figure}[!htb]
\centering
\includegraphics[width=0.48\textwidth]{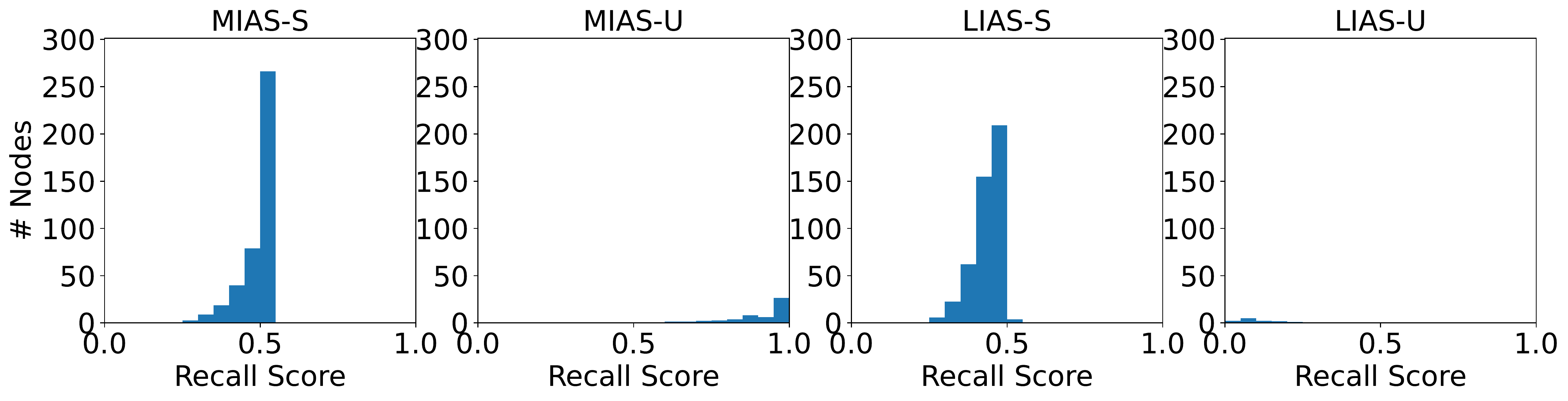}
\caption{\label{fig:graph_evaluate_gcn_cora}Histogram of recall scores of GCN model on Cora dataset with extended masks ($S$ means the nodes are successfully misclassified into the target label, $U$ means not).}
\end{figure}

In contrast, we surprisingly see that for the GCN model on the Cora dataset with MIAS, the unsuccessfully misclassified nodes also have a high recall score as the successfully misclassified nodes. 
%However, by comparing Table~\ref{Table:bkd_graph_domain} and Figure~\ref{fig:cora-gcn} we can see that for GCN model on Cora dataset, although most poisoned testing samples have high recall score under MIA strategy similar to LIA, the ASR of MIA strategy is significantly lower than the LIA strategy, which is not consistent with other models and datasets.  
We assume that the main reason behind this is that, under the MIAS, the feature trigger is injected into the positions of the most representative features. Thus the backdoored model will recognize not only the trigger pattern but also the representative feature pattern for the original label. 
Therefore, for MIAS, it is possible that even the poisoned testing samples that are not successfully misclassified into the target class will have a high recall score. 
We verify this hypothesis by extending the target masks and predicted masks twice the feature trigger length, i.e., $2*n$, and computing the recall scores again.~\footnote{Here, we select an extension rate of $2$. To verify the hypothesis, the extension rate can be set to $\gamma > 1$, and the recall scores of the successfully misclassified nodes are expected to reduce to $1/\gamma$ of that without extended masks.}
The histogram of the new recall scores of the GCN model on the Cora dataset is shown in Fig.~\ref{fig:graph_evaluate_gcn_cora}. 
We also checked the prediction of the backdoored model over the unsuccessfully misclassified nodes. The output indicates that all these nodes are classified into their original classes. 
Comparing Fig.~\ref{fig:cora-gcn} and~\ref{fig:graph_evaluate_gcn_cora}, we observe that the recall scores of the successfully misclassified nodes generally reduce to half of that without extended masks. We believe this is because, for these nodes, the backdoored model recognizes the trigger location exactly, and when we extended the masks twice the trigger length, only half of the features can be recalled. 
However, we can also see that for the MIAS, the recall scores of the unsuccessfully classified nodes are still as high as those without the extended masks. 
This is because the backdoored model recognizes the feature pattern for the original label (that is why these nodes are classified into the original class and the attack is not successful), so even if the masks are extended, the recall score is still high. 
%We verify this hypothesis by splitting the poisoned testing samples into two parts, one is misclassified into the target class successfully and the other one is not, and computing the recall scores for these two part. The histogram of recall scores of these two parts for GCN model on Cora dataset is shown in Figure~\ref{fig:graph_evaluate_gcn_cora}. It can be observed that for MIA strategy, the recall scores for all nodes that are not successfully misclassified into the target class are more than $0.4$ and can be high up to $1.0$ while for LIA, they are all small, i.e., less than $0.4$. 

\section{Related Work}
\label{sec:related}

\subsection{Backdoor Attacks in GNNs}
Several works have conducted backdoor attacks in GNNs. Zhang et al. presented a subgraph-based backdoor attack in GNNs for graph classification task~\cite{zhang2021backdoor}. Xi et al. proposed a subgraph-based backdoor attack in GNNs, for both node classification and graph classification tasks~\cite{xi2021graph}. Xu et al. explored the trigger-injecting position for the graph backdoor attack~\cite{xu2021explainability}, representing the most related work to our paper. However, in that paper, the authors only provided assumptions about the results, and no experimental analysis was given to confirm the assumptions. In this work, we give an empirical analysis of the attack results, which leads to a further understanding of the backdoor attack behavior in GNNs. %\todo{Are there more differences? Is that one difference enough for selling the paper?}
%there is no in-depth analysis and explanation of the attack performance, particularly quantitative analysis.  

\subsection{Explanability in GNNs}
GNNs have become increasingly popular since many real-world data can be naturally represented as graphs, such as social networks, chemical molecules, and financial data~\cite{zhang2020deep, hamilton2020graph}. Consequently, numerous approaches are proposed to explain the predictions of GNNs.
Generally, these methods can be categorized into two mainstream lines of research. One is the parametric explanation methods that are widely used nowadays. For instance, GNNExplainer~\cite{ying2019gnnexplainer} learns soft masks for edges and node features to explain the predictions via mask optimization. The soft masks are randomly initialized and treated as trainable variables.
~\cite{luo2020parameterized} proposed PGExplainer to collectively explains multiple instances with a probabilistic graph generative model. XGNN~\cite{yuan2020xgnn} uses a graph generator to generate class-wise graph patterns to explain GNNs for each class. Vu et al. proposed PGM-Explainer, a Bayesian network on the pairs of graph perturbations and prediction changes~\cite{vu2020pgm}. 

The other line is the non-parametric explanation methods, which do not involve any additional trainable models. They employ heuristics like gradient-like scores obtained by backpropagation as the feature contributions of a specific instance~\cite{baldassarre2019explainability, pope2019explainability, schnake2021higher}

\subsection{Explainability for Backdoor Attacks}
With the thriving development of explainability techniques in machine learning, the attacker can use model explanations to gain knowledge about the model to perform the adversarial attacks~\cite{nadeem2022sok}.
Kuppa et al.~\cite{kuppa2021adversarial} used counterfactual explanations to find the malware features that most heavily impact the classifier decision. They used this knowledge to craft adversarial training samples that efficiently poison the model. Severi et al.~\cite{severi2021explanation} used SHAP to craft backdoor triggers in malware detectors. Utilizing the explanation, they determined which features to poison, resulting in a success rate of up to three times higher than that of a greedy algorithm that does not use explainable artificial intelligence (XAI). Xu et al.~\cite{xu2021explainability} injected backdoors into GNNs by leveraging XAI techniques. While there has been an increasing number of studies on utilizing explanation techniques to implement backdoor attacks in deep learning models, there has been no research on using explanation tools to clarify the backdoor attack behavior in the graph domain.

\section{Conclusion and Future Work}
\label{sec:conclusions}

This paper presents a comprehensive analysis and explanation of graph backdoor attacks with two trigger-injecting strategies; MIAS and LIAS. 
We investigate the node classification task and compare the attack performance for these two strategies. Our findings show that LIAS always achieves higher attack performance than MIAS. 
We further explain the difference with quantitative analysis, which contributes to a further understanding of the backdoor attack behavior in GNNs. 
Future work will include explaining the backdoor attack behavior of two trigger-injecting strategies in the graph classification task. 
More precisely, we would compute the similarity between the new representative subgraph and the old one by calculating the recall score of the target mask and the predicted mask. 

%\todo{we need to better explain relevance, unique challenges, and differentiation from previous works. Also, the design choices}

\bibliographystyle{plain}
\bibliography{references.bib}

\begin{thebibliography}{10}

\bibitem{baldassarre2019explainability}
Federico Baldassarre and Hossein Azizpour.
\newblock Explainability techniques for graph convolutional networks.
\newblock {\em arXiv preprint arXiv:1905.13686}, 2019.

\bibitem{dai2022comprehensive}
Enyan Dai, Tianxiang Zhao, Huaisheng Zhu, Junjie Xu, Zhimeng Guo, Hui Liu,
  Jiliang Tang, and Suhang Wang.
\newblock A comprehensive survey on trustworthy graph neural networks: Privacy,
  robustness, fairness, and explainability.
\newblock {\em arXiv preprint arXiv:2204.08570}, 2022.

\bibitem{errica2019fair}
Federico Errica, Marco Podda, Davide Bacciu, and Alessio Micheli.
\newblock A fair comparison of graph neural networks for graph classification.
\newblock {\em arXiv preprint arXiv:1912.09893}, 2019.

\bibitem{gu2019badnets}
Tianyu Gu, Kang Liu, Brendan Dolan-Gavitt, and Siddharth Garg.
\newblock Badnets: Evaluating backdooring attacks on deep neural networks.
\newblock {\em IEEE Access}, 7:47230--47244, 2019.

\bibitem{hamilton2020graph}
William~L Hamilton.
\newblock Graph representation learning.
\newblock {\em Synthesis Lectures on Artifical Intelligence and Machine
  Learning}, 14(3):1--159, 2020.

\bibitem{hu2015visualizing}
Yifan Hu and Lei Shi.
\newblock Visualizing large graphs.
\newblock {\em Wiley Interdisciplinary Reviews: Computational Statistics},
  7(2):115--136, 2015.

\bibitem{huang2022auc}
Mengda Huang, Yang Liu, Xiang Ao, Kuan Li, Jianfeng Chi, Jinghua Feng, Hao
  Yang, and Qing He.
\newblock Auc-oriented graph neural network for fraud detection.
\newblock In {\em Proceedings of the ACM Web Conference 2022}, pages
  1311--1321, 2022.

\bibitem{jin2022gnnlens}
Zhihua Jin, Yong Wang, Qianwen Wang, Yao Ming, Tengfei Ma, and Huamin Qu.
\newblock Gnnlens: A visual analytics approach for prediction error diagnosis
  of graph neural networks.
\newblock {\em IEEE Transactions on Visualization and Computer Graphics}, 2022.

\bibitem{kipf2017semi}
Thomas~N. Kipf and Max Welling.
\newblock Semi-supervised classification with graph convolutional networks.
\newblock In {\em ICLR}, 2017.

\bibitem{kuppa2021adversarial}
Aditya Kuppa and Nhien-An Le-Khac.
\newblock Adversarial xai methods in cybersecurity.
\newblock {\em IEEE transactions on information forensics and security},
  16:4924--4938, 2021.

\bibitem{liu2021pick}
Yang Liu, Xiang Ao, Zidi Qin, Jianfeng Chi, Jinghua Feng, Hao Yang, and Qing
  He.
\newblock Pick and choose: a gnn-based imbalanced learning approach for fraud
  detection.
\newblock In {\em Proceedings of the Web Conference 2021}, pages 3168--3177,
  2021.

\bibitem{luo2020parameterized}
Dongsheng Luo, Wei Cheng, Dongkuan Xu, Wenchao Yu, Bo~Zong, Haifeng Chen, and
  Xiang Zhang.
\newblock Parameterized explainer for graph neural network.
\newblock {\em Advances in neural information processing systems}, 2020.

\bibitem{nadeem2022sok}
Azqa Nadeem, Dani{\"e}l Vos, Clinton Cao, Luca Pajola, Simon Dieck, Robert
  Baumgartner, and Sicco Verwer.
\newblock Sok: Explainable machine learning for computer security applications.
\newblock {\em arXiv preprint arXiv:2208.10605}, 2022.

\bibitem{pope2019explainability}
Phillip~E Pope, Soheil Kolouri, Mohammad Rostami, Charles~E Martin, and Heiko
  Hoffmann.
\newblock Explainability methods for graph convolutional neural networks.
\newblock In {\em Proceedings of the IEEE/CVF conference on computer vision and
  pattern recognition}, pages 10772--10781, 2019.

\bibitem{schnake2021higher}
Thomas Schnake, Oliver Eberle, Jonas Lederer, Shinichi Nakajima, Kristof~T
  Sch{\"u}tt, Klaus-Robert M{\"u}ller, and Gr{\'e}goire Montavon.
\newblock Higher-order explanations of graph neural networks via relevant
  walks.
\newblock {\em IEEE transactions on pattern analysis and machine intelligence},
  44(11):7581--7596, 2021.

\bibitem{selvaraju2017grad}
Ramprasaath~R Selvaraju, Michael Cogswell, Abhishek Das, Ramakrishna Vedantam,
  Devi Parikh, and Dhruv Batra.
\newblock Grad-cam: Visual explanations from deep networks via gradient-based
  localization.
\newblock In {\em Proceedings of the IEEE international conference on computer
  vision}, pages 618--626, 2017.

\bibitem{sen2008collective}
Prithviraj Sen, Galileo Namata, Mustafa Bilgic, Lise Getoor, Brian Galligher,
  and Tina Eliassi-Rad.
\newblock Collective classification in network data.
\newblock {\em AI magazine}, 2008.

\bibitem{severi2021explanation}
Giorgio Severi, Jim Meyer, Scott~E Coull, and Alina Oprea.
\newblock Explanation-guided backdoor poisoning attacks against malware
  classifiers.
\newblock In {\em USENIX Security Symposium}, pages 1487--1504, 2021.

\bibitem{clean-label-backdoor-attacks}
Alexander Turner, Dimitris Tsipras, and Aleksander Madry.
\newblock Clean-label backdoor attacks.
\newblock 2018.

\bibitem{velickovic2018graph}
Petar Veli{\v{c}}kovi{\'{c}}, Guillem Cucurull, Arantxa Casanova, Adriana
  Romero, Pietro Li{\`{o}}, and Yoshua Bengio.
\newblock {Graph Attention Networks}.
\newblock {\em ICLR}, 2018.

\bibitem{vu2020pgm}
Minh Vu and My~T Thai.
\newblock Pgm-explainer: Probabilistic graphical model explanations for graph
  neural networks.
\newblock {\em Advances in neural information processing systems},
  33:12225--12235, 2020.

\bibitem{wang2021confident}
Xiao Wang, Hongrui Liu, Chuan Shi, and Cheng Yang.
\newblock Be confident! towards trustworthy graph neural networks via
  confidence calibration.
\newblock {\em Advances in Neural Information Processing Systems},
  34:23768--23779, 2021.

\bibitem{wu2020comprehensive}
Zonghan Wu, Shirui Pan, Fengwen Chen, Guodong Long, Chengqi Zhang, and S~Yu
  Philip.
\newblock A comprehensive survey on graph neural networks.
\newblock {\em IEEE transactions on neural networks and learning systems},
  32(1):4--24, 2020.

\bibitem{xi2021graph}
Zhaohan Xi, Ren Pang, Shouling Ji, and Ting Wang.
\newblock Graph backdoor.
\newblock In {\em 30th USENIX Security Symposium (USENIX Security 21)}, pages
  1523--1540, 2021.

\bibitem{xie2020dba}
Chulin Xie, Keli Huang, Pin-Yu Chen, and Bo~Li.
\newblock Dba: Distributed backdoor attacks against federated learning.
\newblock In {\em International conference on learning representations}, 2020.

\bibitem{xu2021explainability}
Jing Xu, Minhui Xue, and Stjepan Picek.
\newblock Explainability-based backdoor attacks against graph neural networks.
\newblock In {\em Proceedings of the 3rd ACM Workshop on Wireless Security and
  Machine Learning}, 2021.

\bibitem{ying2019gnnexplainer}
Zhitao Ying, Dylan Bourgeois, Jiaxuan You, Marinka Zitnik, and Jure Leskovec.
\newblock Gnnexplainer: Generating explanations for graph neural networks.
\newblock {\em Advances in neural information processing systems}, 32, 2019.

\bibitem{yuan2020xgnn}
Hao Yuan, Jiliang Tang, Xia Hu, and Shuiwang Ji.
\newblock Xgnn: Towards model-level explanations of graph neural networks.
\newblock In {\em Proceedings of the 26th ACM SIGKDD International Conference
  on Knowledge Discovery \& Data Mining}, pages 430--438, 2020.

\bibitem{yuan2021explainability}
Hao Yuan, Haiyang Yu, Jie Wang, Kang Li, and Shuiwang Ji.
\newblock On explainability of graph neural networks via subgraph explorations.
\newblock In {\em International Conference on Machine Learning}. PMLR, 2021.

\bibitem{zhang2022trustworthy}
He~Zhang, Bang Wu, Xingliang Yuan, Shirui Pan, Hanghang Tong, and Jian Pei.
\newblock Trustworthy graph neural networks: Aspects, methods and trends.
\newblock {\em arXiv preprint arXiv:2205.07424}, 2022.

\bibitem{zhang2021backdoor}
Zaixi Zhang, Jinyuan Jia, Binghui Wang, and Neil~Zhenqiang Gong.
\newblock Backdoor attacks to graph neural networks.
\newblock In {\em Proceedings of the 26th ACM Symposium on Access Control
  Models and Technologies}, pages 15--26, 2021.

\bibitem{zhang2020deep}
Ziwei Zhang, Peng Cui, and Wenwu Zhu.
\newblock Deep learning on graphs: A survey.
\newblock {\em IEEE Transactions on Knowledge and Data Engineering},
  34(1):249--270, 2020.

\bibitem{zhou2020graph}
Jie Zhou, Ganqu Cui, Shengding Hu, Zhengyan Zhang, Cheng Yang, Zhiyuan Liu,
  Lifeng Wang, Changcheng Li, and Maosong Sun.
\newblock Graph neural networks: A review of methods and applications.
\newblock {\em AI open}, 1:57--81, 2020.

\end{thebibliography}

\end{document}